\documentclass[runningheads]{llncs}
\usepackage{graphicx}
\usepackage{booktabs}
\usepackage{amsmath}
\usepackage{subfig}
\usepackage{hyperref}

\usepackage[]{algorithm2e}

\usepackage{xcolor} 

\usepackage{siunitx}

\usepackage{tabularx}

\begin{document}
\title{RGB-D Railway Platform Monitoring and Scene Understanding for Enhanced Passenger Safety}
\titlerunning{RGB-D Railway Platform Monitoring and Scene Understanding}
%
\author{Marco Wallner\inst{1,2} \and
Daniel Steininger\inst{1,2} \and
Verena Widhalm\inst{1,2} \and \\ Matthias Schörghuber\inst{1,2} \and Csaba Beleznai\inst{1,2}}
\authorrunning{M. Wallner et al.}
%
\institute{Austrian Institute of Technology \and
\email{first.lastname@ait.ac.at}\\
\url{https://www.ait.ac.at}}
\maketitle              

\begin{abstract}
    Automated monitoring and analysis of passenger movement in safety-critical parts of transport infrastructures represent a relevant visual surveillance task. Recent breakthroughs in visual representation learning and spatial sensing opened up new possibilities for detecting and tracking humans and objects within a 3D spatial context. This paper proposes a flexible analysis scheme and a thorough evaluation of various processing pipelines to detect and track humans on a ground plane, calibrated automatically via stereo depth and pedestrian detection. We consider multiple combinations within a set of RGB- and depth-based detection and tracking modalities. We exploit the modular concepts of Meshroom~\cite{Meshroom} and demonstrate its use as a generic vision processing pipeline and scalable evaluation framework.
    Furthermore, we introduce a novel open RGB-D railway platform dataset with annotations to support research activities in automated RGB-D surveillance. We present quantitative results for multiple object detection and tracking for various algorithmic combinations on our dataset. Results indicate that the combined use of depth-based spatial information and learned representations yields substantially enhanced detection and tracking accuracies. As demonstrated, these enhancements are especially pronounced in adverse situations when occlusions and objects not captured by learned representations are present.     
    
    \keywords{RGB-D visual surveillance  \and human detection and tracking  \and tracking evaluation \and evaluation framework \and surveillance dataset}
\end{abstract}
\begin{figure}[ht]
   \centering
   \includegraphics[width=0.95\columnwidth]{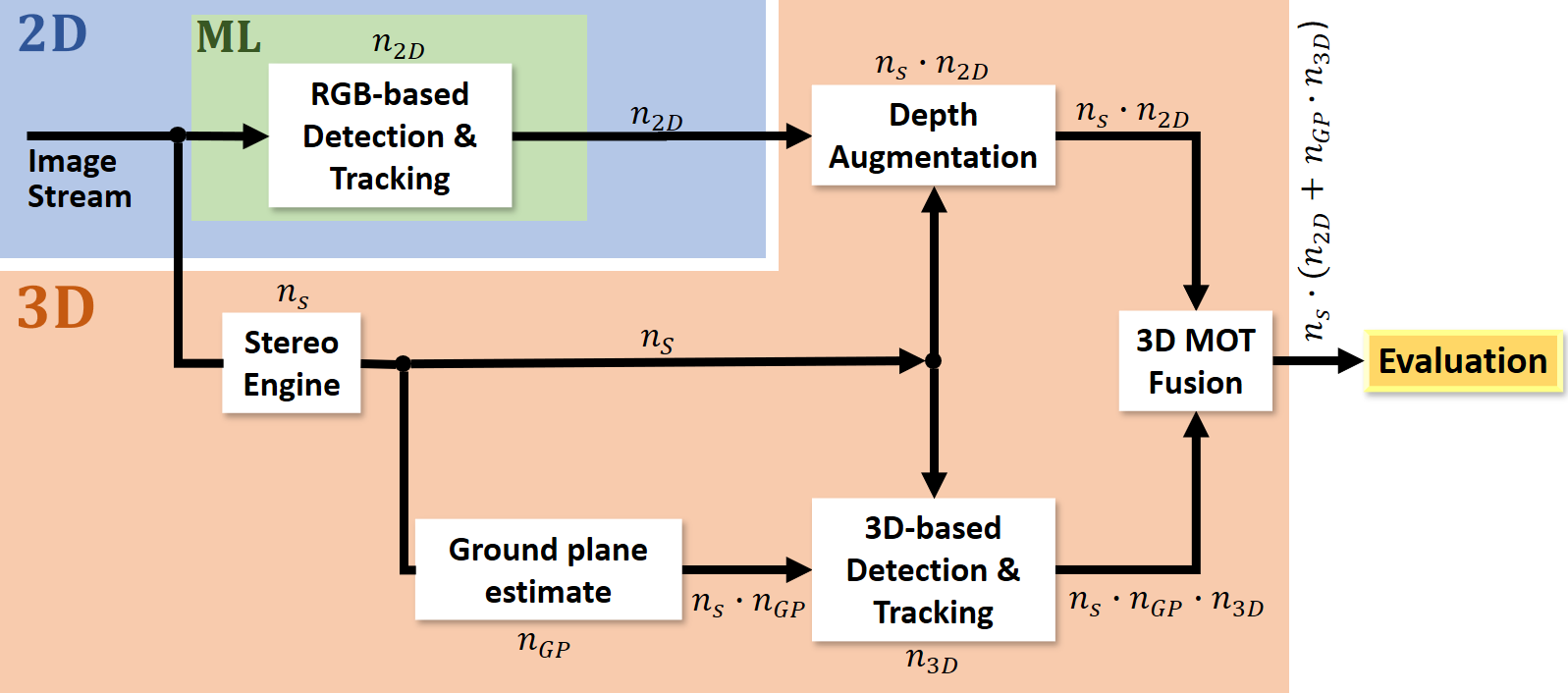}
   \caption{Our combined 2D and 3D analysis concept exploiting learned representations (ML = machine learning) and cues from spatial data. $n_s$, $n_{GP}$, $n_{2D}$ and $n_{3D}$ denote the number of stereo, ground-plane estimation, image-based and depth-based algorithmic methods, along with indication for the number of combinations after each algorithmic module. }
   \label{fig:overview}
\end{figure}

\section{Introduction}
Vision-based robust human detection and tracking are receiving an increasing amount of attention recently. These core technologies are essential enablers for many human-centered applications such as pedestrian safety aspects in autonomous driving~\cite{Combs2019}, public crowd monitoring~\cite{Wang2020}, and human-aware robotics.

With these developments yielding growing algorithmic capabilities, new sets of public benchmarks with added tasks, complexities, and sizes have been proposed. Typical examples for multiple human tracking in RGB images are the MOTChallenge~\cite{MOTChallenge20} and its recent MOTS~\cite{MOTS20} extension. The progress in these benchmarks reflects the recent enhancement of discriminative power introduced by representations via deep learning. However, the use of depth data in the form of RGB-D surveillance has received comparatively less attention. This relative scarcity of combined RGB-D analysis is mainly due to the (i) additional need for a depth sensor or stereo configuration, (ii) the increased computational demand when computing stereo depth, and (iii) the need to create a common representation (e.g., a common ground plane) for combining RGB and depth data.

In this paper, we propose a modular RGB-D processing pipeline (see Figure~\ref{fig:overview}), which allows for the exploration of several RGB-D combined processing schemes. Depth information allows for mapping RGB-based detection results into a 3D space, yielding 3D bounding boxes, which convey size, spatial ordering, and directional information on the ground plane, when being part of a tracking process. Therefore, such information is highly complementary to learned 2D detection results and opens up possibilities to achieve better target separation, segmentation, and occlusion reasoning. In addition, detecting objects from depth data based on geometric cues might yield object proposals that are not captured by learned representations. For example, a box-like object on a platform, when this object type is not part of any class in a training set, will be ignored by learned detectors.
\begin{figure}[ht]
   \centering
   \includegraphics[width=0.95\columnwidth]{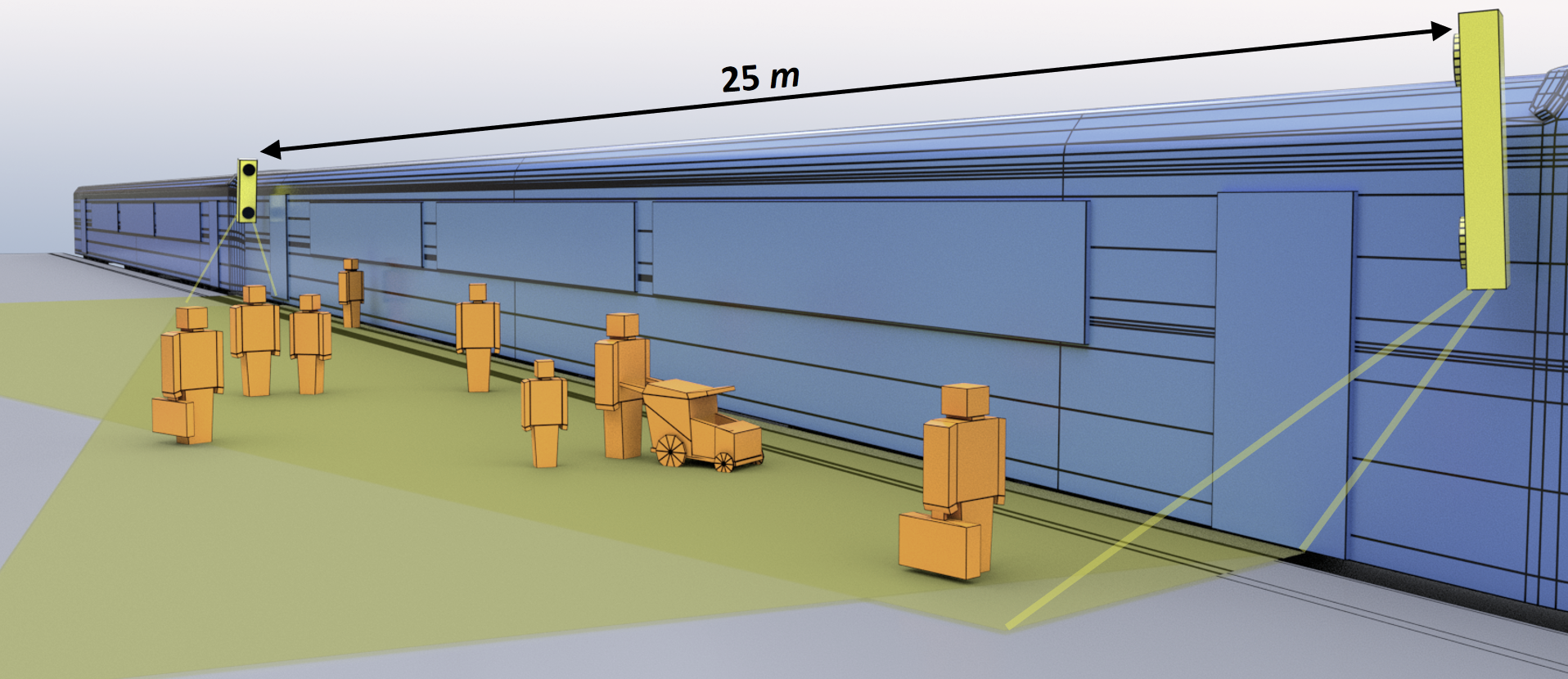}
   \caption{Our dual stereo-camera configuration mounted on a train, exhibiting partially overlapping field-of-views (stereo-cameras are enlarged for highlighting their location and pose).}
   \label{fig:hw_setup}
\end{figure}
This paper introduces the following contributions in the context of RGB-D pedestrian tracking: we present an efficient, sparse-flow based tracking algorithm with an implicit target association step. We present a fusion scheme to combine 2D and depth-based detection results on a common ground plane (see Figure~\ref{fig:hw_setup}). We present a practically interesting aspect via modularizing algorithmic units within Meshroom~\cite{Meshroom} and its use as a generic vision processing pipeline and scalable evaluation framework. We use this modular framework to elaborate and evaluate multiple RGB-D detection- and tracking-schemes. Finally, we introduce RailEye3D, a novel RGB-D railway platform dataset\footnote{\url{https://github.com/raileye3d/raileye3d_dataset}} along with annotations for benchmarking detection and tracking algorithms, to support research activities in the RGB-D surveillance domain. 

\section{Related State of the Art\label{sec:related_sota}} 
Over the last two decades, pedestrian detection and tracking have received much research interest. Their operational domain has been extended towards increasingly complex scenarios. In this process, the following representations have played a key role:

\textbf{Appearance:} Recent years have demonstrated a marked shift from hand-crafted representations to end-to-end learned recognition concepts employing deep distributed representations. The Integral Channel Features~\cite{Dollar2009}, DPM~\cite{Felzenszwalb2008}, and LDCF~\cite{Nam2014} detectors represented well-performing detection algorithms using hand-crafted representations. Modern detection schemes based on deep learning have reached great improvements on common benchmarks, however, at the expense of significantly higher computational costs. A representative example for this computing paradigm is the R-CNN framework~\cite{Zhang2016} and its variants~\cite{Ren2017},~\cite{Zhou2018}. Latest representational concepts such as one-stage inference~\cite{zhou2020tracking},~\cite{liu2020center} of various pedestrian attributes also produce strong performance on common benchmark datasets~\cite{MSCOCO2015},~\cite{zhang2017citypersons},~\cite{Braun2019}. Increasing the data diversity during training~\cite{hasan2020pedestrian} reduces the miss rate on difficult (small size, occlusion) data significantly.

Along with the rise of algorithmic capabilities, larger and more complex pedestrian detection datasets have been proposed. The CityPerson~\cite{zhang2017citypersons}, the EuroCityPerson~\cite{Braun2019}, and the Wider Pedestrian~\cite{WiderPed} datasets are relevant benchmarks indicating the progress of appearance-based pedestrian detection.

\textbf{RGB-D:} Information in depth-data inherently offers ways to lower the ambiguity associated with occlusions and depth ordering. The number of related works is significantly fewer than for appearance-based learned representations. Early RGB-D approaches~\cite{Bertozzi2005} tried to use the depth cue as a segmentation cue in the image space. Approaches with increasing sophistication~\cite{Munoz-Salinas2009},~\cite{Jafari2014} later employed the popular \textit{occupancy map} concept or the voxel space~\cite{Munaro2012} to delineate individual human candidates. The idea of combining the representational strength of learning on combined RGB-D inputs has been proposed by several papers~\cite{Zhou2017},~\cite{Ophoff2019},~\cite{Beleznai2020}. Nevertheless, accomplished improvements are rather small.

Only a few datasets for RGB-D human detection exist. The EPFL RGB-D pedestrian dataset~\cite{Bagautdinov15},~\cite{Ophoff2019} represent relatively simple indoor (lab, corridor) scenarios. The RGB-D People Dataset~\cite{Spinello2011} is an indoor dataset based on Kinect sensor data, depicting limited variations in its scenes. The Kitti Vision Benchmark Suite~\cite{Geiger2013} also contains an RGB-D pedestrian detection task; nevertheless, for a moving platform. Therefore we think that the proposed RailEye3D dataset fills a missing gap within this research domain.

\textbf{Multiple target tracking:} 
Multiple Object Tracking (MOT) is a crucial step to assign detection results to coherent motion trajectories over time. This fundamental temporal grouping task's relevance is reflected by the vast number of tracking methodologies proposed. For example, the MOT benchmark and its evolutions~\cite{Leal-Taixe2017TrackingTracking} display significant progress in scenes of increasing complexity. The multiple object tracking task encompasses the critical aspects of data association, motion estimation, and target representation, each impacting the overall quality. Data association has been intensively investigated and often posed as a classical optimization problem~\cite{Pirsiavash2011},~\cite{Leal-Taixe2011}. Data association has recently been embedded into end-to-end learning schemes, such as predicting future offsets~\cite{zhou2020tracking} or inferring association states~\cite{Braso2020}. Another key aspect is the target appearance representation, where learning representations of great discriminative power produced well-performing tracking schemes~\cite{Wang2019TowardsTracking},~\cite{Zhang2020ATracking}. Accurate target segmentation also substantially contributes to the target representational quality, and it is the main scope of the recent MOTS20 benchmark~\cite{MOTS20}.

\section{Proposed Methodologies and Systemic Concept\label{sec:methods}} 
In this section, we describe our overall processing scheme and its components.
\\ \textbf{Scalable processing and evaluation via Meshroom:} Meshroom~\cite{Meshroom}, as the result of an open-source EU project, offers intuitive node-graph-based modular concepts for off-line vision processing. The framework can be used not only for photogrammetry but also for arbitrary algorithmic units in various implementation forms, such as Python and C++. Its ease-of-use and its capability to create and re-configure a large number of algorithmic pipelines allowed us to define a broad set of configuration, detection, tracking, and evaluation components. 
Following algorithmic blocks have been created: two stereo matching algorithms, multiple 2D and 3D pedestrian detection schemes, a robust-fitting-based ground plane estimation module, multiple tracking methods, a depth augmentation block transforming 2D bounding boxes into 3D by exploiting depth and ground plane information, and a fusion scheme combining multiple 3D observations on the ground plane. A MOT challenge~\cite{Leal-Taixe2017TrackingTracking} evaluation module was also implemented to evaluate individual workflows. Additionally, modules handling various inputs, parameter configurations, visualizations and format conversions complement the algorithmic units. The Meshroom-based framework also offers internal caching of intermediate results, thus allowing for partial tree updates.  
\\ \textbf{Stereo camera setup and depth computation:} We use a binocular stereo setup with a baseline of \SI{400}{\milli\meter} between the two cameras. The RGB cameras are board-level industrial cameras of 2 Megapixels. During rectification, the images are resampled to 1088$\times$1920 pixels with 8 bit quantization. The stereo matching process outputs disparity data alongside with rectified intensity images, congruent to the disparity image. We integrated two stereo matching methods into our processing pipeline, the S3E~\cite{Humenberger2010} and libSGM (based on~\cite{Hirschmuller2008}) techniques. S3E employs a pyramidal implementation of a Census-based stereo matching algorithm, while libSGM combines the Census transform with a semi-global matching step to establish correspondences.
\\ \textbf{Platform detection and automated ground-plane estimation:} For all experiment scenarios, we mounted two stereo camera units with largely overlapping field-of-views on the side of a train. The ground plane homography for a given unit is estimated in a fully automated manner. First, during a calibration phase, we determine image regions representing "walkable surfaces" by aggregating the results of a pedestrian detector~\cite{wu2019detectron2} in the image space and in time. Next, we use a RANSAC-based plane fitting on the depth data to recover the platform's 3D plane parameters. This fitting step is limited to 3D points lying within the previously determined "walkable" areas, and it is also guided by prior knowledge on the camera mounting height, its proximity to the train, and the maximum platform width. This auto-calibration scheme has proved to be successful in all of our scenarios and allowed us to map image-based (denoted as 2D later on) detection results onto a common 3D ground plane. For 3D object detection, such a calibrated ground plane was also a prerequisite for computing an occupancy map representation.     

\subsection{3D Multi-Object Detection and Tracking\label{subsec:mot_solutions}}
To detect human candidates in the depth data, we employ an occupancy map clustering scheme. In the occupancy map, clusters corresponding to humans and compact objects are delineated using a hierarchically-structured tree of learned shape templates~\cite{Beleznai2020}. Thus, local grouping within the two-dimensional occupancy map generates consistent object hypotheses and suppresses background clutter and noise. The detection results are tracked with a conventional frame-to-frame tracking scheme. We employ a standard Bayesian filter based multi-target tracking method where data association is performed by the Hungarian algorithm and states are estimated by a Kalman filter. The target states are represented metric coordinates and velocities, given the calibrated ground plane.

\subsection{2D Object Detection and Tracking Schemes}
\subsubsection{FOT /Fast Object Tracker/:}
We included our own tracking framework called \textit{Fast Object Tracker} (FOT) for the experiments and evaluations. This tracking algorithm has been developed to be detector-agnostic and to exploit run-time optimized sparse optical flow computation for the data association step. Sparse optical flow estimation relies on a high number of local features with distinctive and compact descriptor representations. Matching is performed within a larger temporal neighborhood, resulting in an enhanced discovery of coherently moving object points and suppression of spurious correspondences. For stationary camera setups, background modeling can be computed as an additional hint for identifying potential tracking targets, which along with the sparse-flow computation, is parallelized on the GPU to provide real-time performance even on embedded devices. Our experiments used a pre-trained YOLOv3 \cite{redmon2018yolov3} detector for generating object proposals to be tracked.
\\ \textbf{Third-Party Tracking Schemes:}
In addition to the FOT-tracker, we integrated two recent state-of-the-art tracking schemes employing highly discriminative target representations for re-identification:
\\ \textbf{TRT-MOT /Towards Realtime MOT/}~\cite{Wang2019TowardsTracking} incorporates an appearance embedding model into a single-shot detector. This integration is performed by formulating the learning task with multiple objectives: anchor classification, bounding box regression, and learning a target-specific embedding. The representational homogeneity and joint optimization for these distinct tasks yield an enhanced detection accuracy and target matching capability.
\\ \textbf{FairMOT}~\cite{Zhang2020ATracking} builds upon TRT-MOT, but proposes an anchor-free approach. Multiple nearby anchors in TRT-MOT belonging to the same target might lower the representational specificity, a deficiency removed in FairMOT.
Due to this algorithmic enhancement, FairMOT outperforms TRT-MOT in several benchmarks.

\subsection{Fusion of MOT Results}
The variety of proposed multiple object tracking concepts have increased in recent years. These algorithms often approach the tracking task with different strategies, or rely on different input modalities. The presented fusion scheme's key idea is to conceive a general framework to fuse tracking results independent of their origin. To allow 2D and 3D approaches to be fused, image-based (2D) tracking results are augmented with depth information to create a common 3D bounding box representation.
\\ \textbf{Depth Estimates:} Most recent tracking solutions perform object detection and tracking on a 2D bounding box basis (see Section~\ref{sec:related_sota}). Estimating these objects' depth as the mean of all measurements inside the bounding box leads to wrong results in general as data points, which are part of the background, are used too. The structure of most objects can be exploited to overcome this problem without using semantic segmentation solutions. The objects are most likely located in the middle of the detected bounding box with no holes in their centers.
The depth measurements are weighted with a 2D Gaussian kernel centered in the bounding box's middle to exploit this. 
The $p=2$ dimensional kernel can be defined in general as
\begin{equation}
    f\left(\mathbf{x}, \mathbf{\mu}, \mathbf{\Sigma} \right) = \frac{1}{\sqrt{\left(2 \pi\right)^p \cdot Det\left(\mathbf{\Sigma}\right)}} \exp\left( -\frac{1}{2} \left( \mathbf{x} - \mathbf{\mu} \right)^{T} \mathbf{\Sigma}^{-1} \left( \mathbf{x} - \mathbf{\mu} \right) \right)
\end{equation}
with $\mathbf{x} = \begin{pmatrix}u && v\end{pmatrix}^{T}$, $\mathbf{\mu} = \begin{pmatrix}c_u && c_v\end{pmatrix}^{T}$ and $\mathbf{\Sigma} = \begin{pmatrix}\sigma_u && 0 \\ 0 && \sigma_v \end{pmatrix}$. $c_u$ and $c_v$ are chosen to be half of the bounding box width and height, $\sigma_u$ and $\sigma_v$ to the size of the bounding box. $u$ and $v$ are the column and row indices inside the bounding box.
For our use case, this can be simplified with the width and height of the bounding box $w_{BB}$ and $h_{BB}$ to:
\begin{equation}
    w(u, v) = \frac{1}{2\pi\sqrt{w_{BB} h_{BB}}} \exp \left( - \left( \frac{\left(u-\frac{w_{BB}}{2}\right)^2}{2 w_{BB}^2}  + \frac{\left(v-\frac{h_{BB}}{2}\right)^2}{2 h_{BB}^2} \right)  \right) 
\end{equation}
As there is not always a valid depth measurement for all pixels, a validity function $\phi$ is introduced as
\begin{equation}
        \phi(u,v)= 
\begin{cases}
    1,& \text{if } d(u,v) \text{ is valid}\\
    0,              & \text{otherwise.}
\end{cases}
\end{equation}
The normalizing weight sum $W_i$ for the $i^{th}$ bounding box can be calculated as
\begin{equation}
    W_i = \sum_{u=0}^{w_{BB}-1} \sum_{v=0}^{h_{BB}-1} \phi(u,v) \cdot w(u,v).
\end{equation}
With this, the depth $d_{i_{est}}$ is estimated as
\begin{equation}
    d_{i_{est}} = \frac{1}{W_i} \sum_{u=0}^{w_{BB}-1} \sum_{v=0}^{h_{BB}-1} \phi(u,v) \cdot w(u, v) \cdot d(u, v).
\end{equation}
This calculated distance can be interpreted as the distance to the object's surface facing the camera. With this, the object's 3D volume is estimated to have the same depth as its width. Resulting 3D bounding boxes are finally transformed into a common reference coordinate system (using the estimated platform's plane homography) to allow fusion of multiple camera views.
\\ \textbf{Fusion Algorithm:} Having these multiple tracking results of the same scene as 3D bounding boxes in a common reference frame alleviates the fusion problem. The abstract approach is described in Algorithm~\ref{algo:new_observation} and \ref{algo:check_contained}.
The main idea is to treat the incoming tracker results as tracklets per object. These tracklets are fused into existing tracks (with a new consistent track id) if their current 3D bounding box intersects enough (IoU threshold) or are fully enclosed (i.e., intersection over envelope IoE).

\begin{figure}
\begin{algorithm}[H]
 \KwData{Detection $det$}
 \KwResult{Track-ID consistent fusion of new observations}
 \If{$tracklet\_id$ = DetectionAlreadyInTrackletHistory($det$)}
 {addNewDetectionToTracklet($det$, $tracklet\_id$)\;
  return\;}
  $IoU$, $IoE$, $tracklet\_id$ = getTrackletWithMostOverlap($det$)\;
  \eIf{$IoU$ $\geq$ $thres$ or $IoE$ $\geq$ $thres$}
  {fuseDetectionWithTracklet($det$, $tracklet\_id$)\;}
  {setupNewTracklet($det$)\;}
   \caption{Fuse new observations (tracklet) with consistent tracks.\label{algo:new_observation}}
\end{algorithm}
 \end{figure}
\begin{figure}
\begin{algorithm}[H]
 \KwData{Detection $det$, TrackletManager $t\_mgr$}
 \KwResult{True if observations with same tracker ID and track ID already contained}
 \eIf{$\exists$ $t$ $\in$ $t\_mgr$ : $(track\_ID(t) == track\_ID(det)) \quad \wedge $ \\$ \qquad \qquad \qquad \qquad (tracker\_ID(t) == tracker\_ID(det))$}{return $true$\;}{return $false$\;}
\caption{Check if new observation belongs to existing tracklet.\label{algo:check_contained}}
\end{algorithm}
 \end{figure}

\section{The RailEye3D Railway Platform Dataset \label{subsec:data}}
Evaluating the performance of multi-object-tracking algorithms requires domain-specific test data. While many available datasets traditionally focus on persons as their main tracking target \cite{zhang2017citypersons}, \cite{Braun2019}, \cite{WiderPed}, few of them provide RGB-D data ~\cite{Bagautdinov15},~\cite{Ophoff2019}, \cite{Spinello2011}, \cite{Geiger2013}, and none of them consider safety aspects on train platforms as an application setting to the best of our knowledge.  
Therefore, the RailEye3D stereo dataset was developed to include a diverse set of train-platform scenarios for applications targeting passenger safety and assisting the clearance of train dispatching. Image data is captured by two stereo camera systems mounted on the train's sides (Figure \ref{fig:hw_setup}) with partially overlapping fields-of-view, ensuring that objects occluded in one perspective are still visible in the opposing one. Three representative test sequences were selected for creating annotations of adults and children and selected objects relevant to the scenario, such as wheelchairs, buggies, and backpacks. Every $10^{th}$ frame of a sequence was annotated to increase annotation efficiency while facilitating a sufficiently precise evaluation of detection and multi-object-tracking methods. Instance and tracking annotations are created manually in two stages using the open-source annotation tool Scalabel \cite{scalabel}. Each object is labeled with a parameter representing its degree of occlusion (0\%, 25\%, 50\%, 75\%, 100\%), and persons are furthermore assigned a unique ID consistent through all scenes and camera views to enable the evaluation of person tracking and re-identification tasks. The annotations include the position of the safety line defining the area to be cleared before dispatch. The use of stereo cameras additionally provides the opportunity to benchmark algorithms integrating depth cues. Table \ref{table:dataset_1st_batch_statistics} provides an overview of relevant annotation statistics for each scene. 
\begin{table}[ht]
\centering
\setlength{\tabcolsep}{8pt} 
\begin{tabular}{c  c  c  c  c}
\toprule
\textbf{Scene}	& \textbf{\#Frames}	& \textbf{\#Annotated} & \textbf{\#Persons} & \textbf{\#Objects}\\
\midrule
\textbf{1}   & 5.505     & 540      & 8.707      & 3.630\\
\textbf{2}   & 2.999     & 300      & 2.205      & 1.656\\
\textbf{3}   & 3.363     & 338      & 4.721      & 3.951\\
\bottomrule
\end{tabular}
\caption{Annotation overview of the RailEye3D dataset}
\label{table:dataset_1st_batch_statistics}
\end{table}
\begin{figure}[t]
   \centering
   \includegraphics[width=0.85\columnwidth]{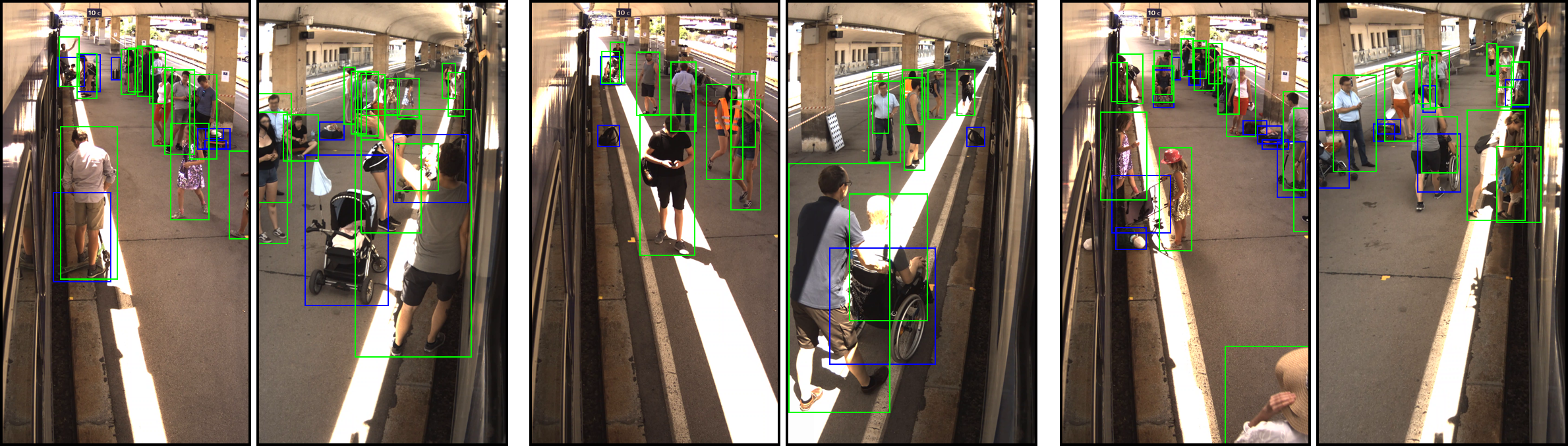}
   \caption{Sample annotations of the RailEye3D dataset for persons and things.}
   \label{fig:dataset_1st_batch}
\end{figure}

\begin{figure}[t]
   \centering
   \includegraphics[width=1.0\columnwidth]{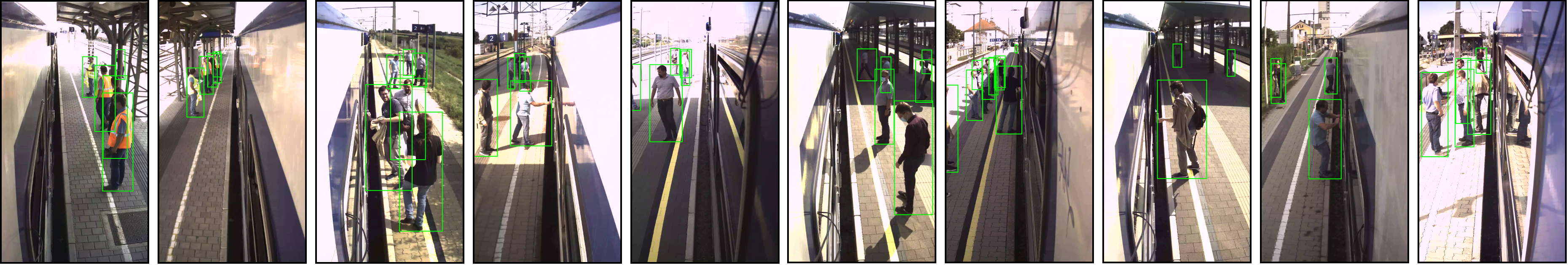}
   \caption{Sample frames of the RailEye3D stereo dataset recorded at 10 different railway platforms.}
   \label{fig:dataset_2nd_batch}
\end{figure}

The selected scenes shown in Figure~\ref{fig:dataset_1st_batch} include typical scenarios such as people waiting and boarding trains, riding scooters and skateboards, or gathering in groups. Children playing and persons using wheelchairs are represented as well, along with some uncommon situations of luggage left on the platform or objects falling between platform and train. The dataset captures a variety of lighting conditions and environmental context (see Figure \ref{fig:dataset_2nd_batch}). 

\section{Results and Discussion}
\begin{figure}[!t]
   \centering
   \includegraphics[width=1.0\columnwidth]{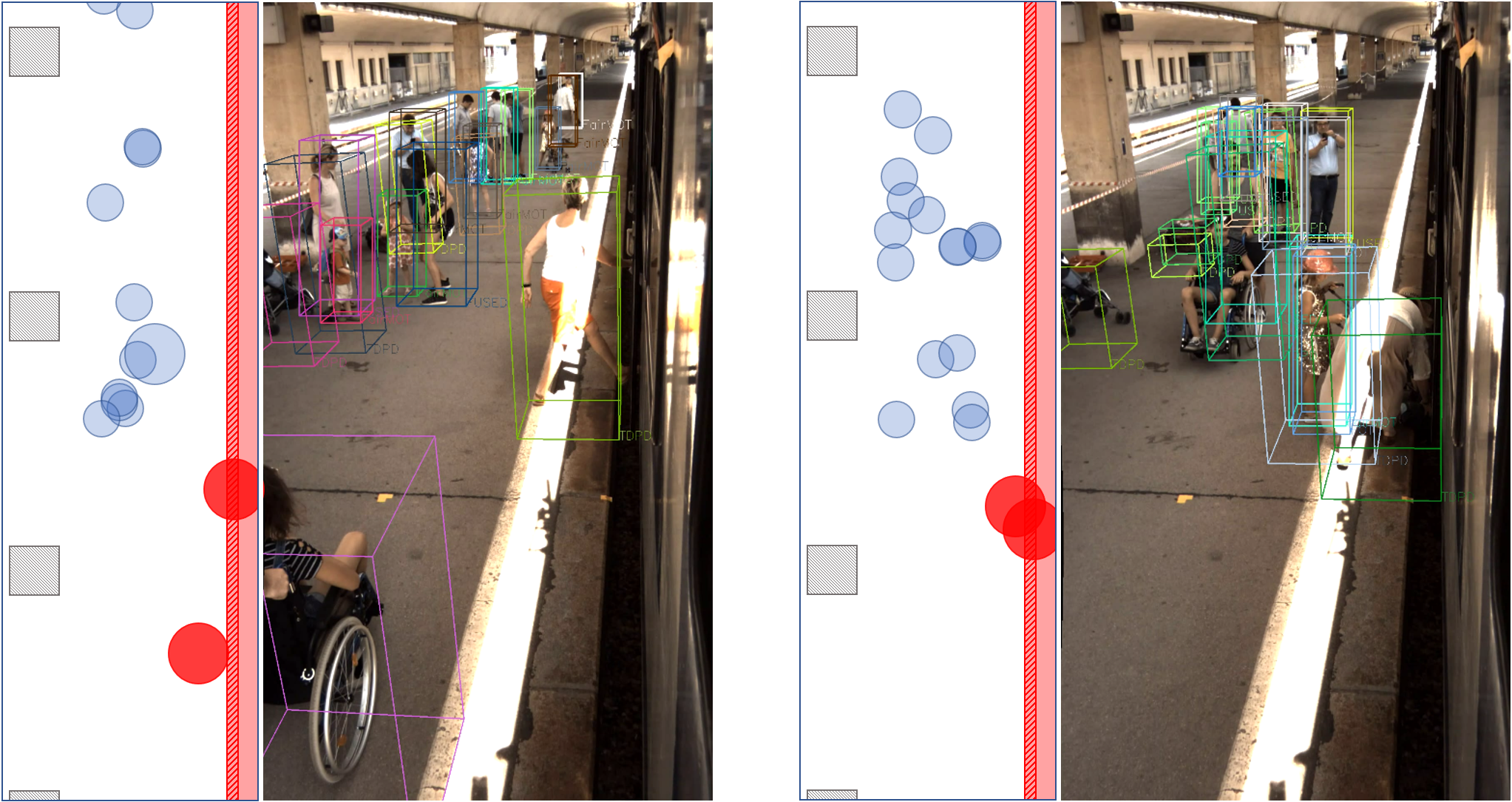}
   \caption{Birds-eye view representations of the platform with critical objects in red and the corresponding fused detection results back-projected into the image.}
   \label{fig:scene_fusion_result}
\end{figure}

Using our presented re-configurable workflow concept, we created and evaluated multiple processing pipelines for the train platform surveillance task. 
\\ \textbf{Selected methods:} The compared stereo engines (see Section~\ref{sec:methods}) provided comparable depth data quality. Therefore, for the sake of brevity, we only present results for one (S3E) stereo depth computation technique. The system shown in Figure~\ref{fig:overview} is evaluated with the following configurations, where $n_s$, $n_{GP}$, $n_{2D}$ and $n_{3D}$ denote the number of stereo, ground-plane estimation, image-based and depth-based algorithmic methods, respectively:
\begin{itemize}
    \item One stereo engine (S3E): $n_s = 1$
    \item One platform estimator (disabled masking, see Section~\ref{sec:methods}): $n_{GP} = 1$
    \item Three image-based (2D) detection and tracking algorithms: $n_{2D} = 3$
    \begin{itemize}
        \item Fast Object Tracker - FOT
        \item Towards Realtime MOT - TRT-MOT
        \item FairMOT
    \end{itemize}
    \item One depth-based detection and tracking algorithm: $n_{3D} = 1$
\end{itemize}
This leads to a fusion of $n = n_s \cdot (n_{2D} + n_{GP} \cdot n_{3D}) = 4$ different MOT results and all their combinations $c = \sum_{k = 1}^{n} \binom{n}{k} = 15$ for each scene.
\\ \textbf{Selected evaluation metrics:} Our evaluations use established metrics for object detection and tracking (see Table~\ref{tab:mot_metrics_definitions}), as defined in  \cite{Bernardin2006}, \cite{Li2009}, and \cite{Milan2016}:
\begin{table*}
    \centering
    \small
\begin{tabularx}{\columnwidth}{lX}
\toprule
    \textbf{IDF1} & Global min-cost F1 score. \\
    \textbf{IDP} & Global min-cost precision. \\
    \textbf{IDR} & Global min-cost recall. \\
    \textbf{Rcll} & Number of correct detections over the total number of ground-truth objects. \\
    \textbf{Prcn} &Number of correct detections over the total number of detected objects. \\
    \textbf{GT} & Sum of tracks in the ground-truth. \\
    \textbf{MT} & Number of objects tracked for at least 80\% of their lifespan.\\
    \textbf{PT} & Number of objects tracked between 20\% and 80\% of their lifespan. \\
    \textbf{ML} & Number of objects tracked less than 20\% of their lifespan. \\
    \textbf{FP} &  Number of detected objects without corresponding ground-truth source. \\
    \textbf{FN} & Number of missed detections. \\
    \textbf{IDs} & Sum of identity switches of tracked objects.\\
    \textbf{FM} & Total number of track fragmentations. \\
    \textbf{MOTA} & Incorporates tracking errors including false positives, misses and mismatches. \\
    \textbf{MOTP} & Relating the average positional error across all correctly tracked objects.\\
\bottomrule
\end{tabularx}

    \caption{Multiple object tracking metric definitions.}
    \label{tab:mot_metrics_definitions}
\end{table*}
\\ \textbf{Task settings:} Given on the annotation diversity of our test scenes, we defined two distinct task settings for evaluating algorithmic workflows. The task setting \textit{ALL} covers the detection and tracking of all annotated humans and objects on the platform. The task setting \textit{PEDS} focuses only on humans with at least 25\% visibility. These split task definitions help us to characterize the individual workflows in multiple terms. The influence of input data quality such as the depth quality for small, thin or flat objects negatively impacts results in the \textit{ALL} setting. Similarly, objects not covered by learned representations in the \textit{ALL}-Set will likely contribute to high miss rates (FN) or lower tracking rates (ML).
\\ \textbf{Experiments and results:} We evaluated our algorithmic workflows on 3 different scenes of the proposed dataset (Section~\ref{subsec:data}). Evaluation results for the tasks settings \textit{ALL} and \textit{PEDS} are shown in the Tables~\ref{tab:eval_combined} and~\ref{tab:eval_combined_pers_only_no_occ}, respectively. As it can be seen from the tables, for both task settings, the combination between algorithmic modalities lowers the FN rate, increases the number of mostly-tracked (MT) targets, but accumulates the FP rate. However, we observe that most false positives are of transient nature, implying that a human operator monitoring results, can probably easily discard such spurious signals (see Figure~\ref{fig:scene_fusion_result}).

Experiments in the \textit{ALL} setting (Table~\ref{tab:eval_combined}) clearly indicate the added values of combining depth information with learned representations. Many objects which are not detected by 2D detection schemes, are found by the combined methods, leading to lower FN rates and better MT rates. Such difficult objects, where learned representations are failing and depth-based detection are still yielding valid proposals, are for example pedestrians partially visible or in unusual poses (bending forward), objects not part of learned representations (buggy). On the other hand, learned representations excel in cases when objects are far from the camera and small, or their form is thin (skateboard, luggage, sticks). In such cases depth data contains little evidence for object presence.
\begin{table*}
    \centering
    \tiny
    \begin{tabular}{lrrrrrrrrrrrrrrr}
\toprule
           Algorithm &  IDF1 &   IDP &   IDR &  Rcll &  Prcn &  GT &  MT &  PT &  ML &    FP &    FN &   IDs &    FM &   MOTA &  MOTP \\
\midrule
                 (i) & 45.2\% & 44.6\% & 34.0\% & 18.4\% & 36.5\% &  50 &   0 &  24 &  26 &  3138 &  7899 &   394 &   764 & -17.5\% &  \textbf{ 0.40} \\
                (ii) & 45.2\% & 46.3\% & 33.0\% & 32.0\% & \textbf{71.0\%} &  50 &   4 &  33 &  13 &  \textbf{1279} &  6825 &   845 &   654 &   9.7\% & 0.28 \\
               (iii) & \textbf{55.7\%} & \textbf{55.6\%} & 41.4\% & 34.3\% & 70.8\% &  50 &   9 &  27 &  14 &  1427 &  6544 &   523 &   570 &  14.8\% &  0.31 \\
                (iv) & 54.8\% & \textbf{55.6\%} & 40.1\% & 32.4\% & 70.2\% &  50 &   7 &  28 &  15 &  1441 &  6736 &   \textbf{293} &   \textbf{499} &  \textbf{15.4\%} &  0.30 \\
 (i)+(ii)+(iii)+(iv) & 41.7\% & 31.2\% & \textbf{46.6\%} & \textbf{47.2\%} & 38.3\% &  50 &  \textbf{13} &  30 &   \textbf{7} &  7556 &  \textbf{5157} &  1055 &   942 & -40.3\% &  0.31 \\
      (i)+(ii)+(iii) & 43.0\% & 33.7\% & 43.9\% & 45.5\% & 43.7\% &  50 &  10 &  31 &   9 &  5742 &  5329 &  1078 &  1005 & -24.3\% &  0.31 \\
       (i)+(ii)+(iv) & 43.7\% & 34.9\% & 43.5\% & 44.3\% & 44.9\% &  50 &  12 &  28 &  10 &  5432 &  5497 &   940 &   904 & -19.8\% &  0.30 \\
      (i)+(iii)+(iv) & 45.6\% & 36.2\% & 45.7\% & 41.7\% & 41.7\% &  50 &  10 &  29 &  11 &  5721 &  5660 &   763 &   892 & -24.6\% &  0.32 \\
     (ii)+(iii)+(iv) & 45.3\% & 36.7\% & 44.1\% & 43.1\% & 45.9\% &  50 &  11 &  31 &   8 &  4901 &  5697 &   901 &   697 & -17.7\% &  0.30 \\
            (i)+(ii) & 43.2\% & 36.8\% & 38.8\% & 41.0\% & 51.5\% &  50 &   5 &  33 &  12 &  3789 &  5831 &  1007 &  1002 &  -8.1\% &  0.32 \\
           (i)+(iii) & 47.9\% & 40.4\% & 43.5\% & 39.2\% & 47.9\% &  50 &  10 &  27 &  13 &  4287 &  5932 &   754 &   943 & -10.9\% &  0.33 \\
            (i)+(iv) & 48.0\% & 41.2\% & 42.8\% & 36.9\% & 47.2\% &  50 &   7 &  29 &  14 &  4129 &  6190 &   518 &   852 &  -9.5\% &  0.32 \\
          (ii)+(iii) & 48.4\% & 42.6\% & 41.7\% & 40.7\% & 57.1\% &  50 &   9 &  31 &  10 &  2946 &  5926 &   894 &   718 &   0.2\% &  0.30 \\
           (ii)+(iv) & 49.5\% & 44.5\% & 41.4\% & 39.5\% & 58.8\% &  50 &  11 &  27 &  12 &  2730 &  6092 &   717 &   665 &   4.4\% &  0.28 \\
          (iii)+(iv) & 51.7\% & 46.6\% & 43.0\% & 36.5\% & 55.1\% &  50 &   9 &  28 &  13 &  2834 &  6310 &   560 &   578 &   1.0\% &  0.31 \\
\bottomrule
\end{tabular}

    \caption{Individual and combined algorithmic results for the \textit{ALL} setting, covering all object types. (i) "3D Person Detector and Tracker", (ii) "FOT", (iii) "TRT-MOT", and (iv) "FairMOT".}
    \label{tab:eval_combined}
\end{table*}
\begin{table*}
    \centering
    \tiny
    \begin{tabular}{lrrrrrrrrrrrrrrr}
\toprule
           Algorithm &  IDF1 &   IDP &   IDR &  Rcll &  Prcn &  GT &  MT &  PT &  ML &    FP &    FN &  IDs &   FM &   MOTA &  MOTP \\
\midrule
                 (i) & 18.9\% & 21.0\% & 17.2\% & 41.5\% & 50.7\% &  46 &   5 &  28 &  13 &  2627 &  3742 &  315 &  439 &  -4.4\% &  0.37 \\
                (ii) & 12.4\% & 14.5\% & 11.1\% & 50.2\% & 68.1\% &  46 &   4 &  35 &   7 &  1363 &  3255 &  822 &  554 &  11.6\% &  0.28 \\
               (iii) & 27.7\% & 31.2\% & 25.1\% & 55.3\% & 69.8\% &  46 &  11 &  28 &   7 &  1480 &  2951 &  497 &  507 &  22.9\% &  0.30 \\
                (iv) & 26.2\% & 30.6\% & 23.0\% & 52.7\% & 69.7\% &  46 &   8 &  29 &   9 &  1456 &  3105 &  291 &  473 &  24.3\% &  0.30 \\
 (i)+(ii)+(iii)+(iv) & 15.4\% & 11.5\% & 23.4\% & 59.7\% & 29.3\% &  46 &  12 &  28 &   6 &  8695 &  2682 &  755 &  532 & -97.4\% &  0.38 \\
      (i)+(ii)+(iii) & 16.2\% & 12.7\% & 22.6\% & 60.3\% & 34.0\% &  46 &  10 &  31 &   5 &  7039 &  2637 &  748 &  534 & -69.8\% &  0.38 \\
       (i)+(ii)+(iv) & 16.9\% & 13.5\% & 22.6\% & 57.8\% & 34.5\% &  46 &  10 &  31 &   5 &  6814 &  2751 &  620 &  532 & -62.7\% &  0.38 \\
      (i)+(iii)+(iv) & 19.6\% & 15.7\% & 25.9\% & 57.5\% & 34.8\% &  46 &   8 &  29 &   9 &  6850 &  2901 &  570 &  535 & -60.3\% &  0.38 \\
     (ii)+(iii)+(iv) & 15.3\% & 12.8\% & 19.3\% & 46.8\% & 31.2\% &  46 &   8 &  30 &   8 &  6302 &  3375 &  677 &  530 & -73.9\% &  0.39 \\
            (i)+(ii) & 17.0\% & 14.7\% & 20.3\% & 56.1\% & 41.0\% &  46 &   4 &  36 &   6 &  4875 &  2865 &  723 &  581 & -36.9\% &  0.38 \\
           (i)+(iii) & 19.3\% & 16.5\% & 23.0\% & 55.6\% & 40.0\% &  46 &   7 &  30 &   9 &  5235 &  3018 &  522 &  542 & -36.1\% &  0.38 \\
            (i)+(iv) & 15.1\% & 27.6\% & 15.5\% & 32.3\% & 42.4\% &  46 &   1 &  19 &  26 &  3848 &  4380 &  251 &  346 & -29.0\% &  0.39 \\
          (ii)+(iii) & 15.3\% & 14.3\% & 16.5\% & 44.3\% & 38.5\% &  46 &   7 &  30 &   9 &  4187 &  3530 &  664 &  541 & -41.5\% &  0.39 \\
           (ii)+(iv) & 16.9\% & 16.0\% & 18.0\% & 44.6\% & 39.9\% &  46 &   8 &  28 &  10 &  4012 &  3509 &  564 &  505 & -34.3\% &  0.38 \\
          (iii)+(iv) & 17.8\% & 17.0\% & 18.7\% & 39.4\% & 36.0\% &  46 &   6 &  23 &  17 &  4351 &  3855 &  407 &  424 & -39.4\% &  0.39 \\
\bottomrule
\end{tabular}

    \caption{Individual and combined algorithmic results for the \textit{PEDS} setting, considering pedestrians only, with occlusions less than 75\%. (i) "3D Person Detector and Tracker", (ii) "FOT", (iii) "TRT-MOT", and (iv) "FairMOT".}
    \label{tab:eval_combined_pers_only_no_occ}
\end{table*}

The experiments in the \textit{PEDS} setting (Table~\ref{tab:eval_combined_pers_only_no_occ}) focus on easier criteria (only humans, mostly visible), where one would assume that both depth- and image-based modalities perform well, even when employed independently. Surprisingly, even in this case, the combination of 2D and 3D modalities significantly improves detection and tracking qualities. We explain this improvement by multiple factors: (i) the combination of weak detection responses in 2D and 3D generates more stable detection responses, yielding recall and tracking stability improvements; (ii) the 3D spatial context, such as depth ordering, introduces advantages when performing multiple object tracking. Occlusion handling, when performed explicitly or by implicit means (i.e., better target separability when modeling the state space in 3D), leads to enhanced tracking stability.
In light of all experiment workflows, evaluation results suggest that a combination of 3D pedestrian detection and tracking, combined with the FOT- and FairMOT schemes, generates the best results in both task settings. This combination is indicated by (i)+(ii)+(iii) in the result Tables~\ref{tab:eval_combined} and~\ref{tab:eval_combined_pers_only_no_occ}.

\section{Conclusion and Outlook}
This paper presents the evaluation for a set of algorithmic combinations in terms of multiple object detection and tracking using RGB-D inputs. Such functionalities play a crucial role in accomplishing automated platform surveillance in transport infrastructures. 

We demonstrate a modular processing approach to create and adapt diverse computer vision processing pipelines, a scheme suggesting an impact even beyond the presented scope. We also introduce a novel annotated railway platform dataset, simultaneously recorded by two RGB-D cameras, with largely overlapping fields-of-view. We demonstrate benchmark results on this data using the re-configurable pipeline. 
The presented combination of RGB-D modalities achieves high recall and stable track generation for most scene targets. This virtue of the combined algorithmic schemes renders them apt to use in an automated safety monitoring context, where train operators receive real-time situational feedback on potential passenger safety issues in the train's vicinity.

The proposed dataset and the learned failure modes in the experiments imply many future opportunities. The present paper does not exploit the dual-camera setup, as the analysis is only carried out in one view. Auto-calibration and joint processing of the two views will likely mitigate issues related to occluded and small-sized targets. Furthermore, the fusion of multiple tracking results could be formulated as a spatio-temporal graph-based fusion problem.

\section*{Acknowledgement}
The authors would like to thank both the Federal Ministry for Climate Action, Environment, Energy, Mobility, Innovation and Technology, and the Austrian Research Promotion Agency (FFG) for co-financing the RAIL:EYE3D research project (FFG No. 871520) within the framework of the National Research Development Programme "Mobility of the Future". In addition, we would like to thank our industry partner EYYES GmbH, Martin Prießnitz with the Federal Austrian Railways (ÖBB) for enabling the recordings, and Marlene Glawischnig and Vanessa Klugsberger for support in annotation.

%
%
%
%
%
\bibliographystyle{splncs04}
\bibliography{local}

\end{document}